\documentclass[10pt,twocolumn,letterpaper]{article}
\pdfoutput=1

\usepackage{cvpr}
\usepackage{multirow}
\usepackage{times}
\usepackage{epsfig}
\usepackage{graphicx}
\usepackage{amsmath}
\usepackage{amssymb}
\usepackage{float}
\usepackage{multicol}
\usepackage{authblk}
\usepackage{amssymb}

\usepackage[breaklinks=true,bookmarks=false]{hyperref}

\cvprfinalcopy 

\def\cvprPaperID{****} 

\begin{document}

\title{StRDAN: Synthetic-to-Real Domain Adaptation Network for Vehicle Re-identification}

\author[1]{Sangrok Lee}
\author[1]{Eunsoo Park}
\author[2]{Hongsuk Yi}
\author[3]{Sang Hun Lee\thanks{Corresponding author}}
\affil[1]{MODULABS}
\affil[ ]{srl@modulabs.ai, es.park@modulabs.co.kr}
\affil[2]{Korea Institute of Science and Technology Information}
\affil[ ]{hsyi@kisti.re.kr}
\affil[3]{Kookmin University}
\affil[ ]{shlee@kookmin.ac.kr}

\maketitle
\begin{abstract}
Vehicle re-identification aims to obtain the same vehicles from vehicle images. This is challenging but essential for analyzing and predicting traffic flow in the city. Although deep learning methods have achieved enormous progress for this task, their large data requirement is a critical shortcoming. Therefore, we propose a synthetic-to-real domain adaptation network (StRDAN) framework, which can be trained with inexpensive large-scale synthetic and real data to improve performance. The StRDAN training method combines domain adaptation and semi-supervised learning methods and their associated losses. StRDAN offers significant improvement over the baseline model, which can only be trained using real data, for VeRi and CityFlow-ReID datasets, achieving 3.1\% and 12.9\% improved mean average precision, respectively.

\end{abstract}
\section{Introduction}
Vehicle re-identification (Re-ID) aims to identify the same vehicles that are captured by various cameras. It is an essential technology for analyzing and predicting traffic flow in smart cities and uses visual appearance based Re-ID methods in general. However, vehicle Re-ID is challenging for two reasons. 

\begin{itemize}
\item Different lighting and complex environments create difficulties with appearance based vehicle Re-ID, and large apparent variations can be generated using different cameras.
\item Different vehicles can be visually very similar when they are in the same type category.
\end{itemize}

\begin{figure}[t!]
\begin{center}
\label{fig:fig1}
  \includegraphics[width=1\linewidth]{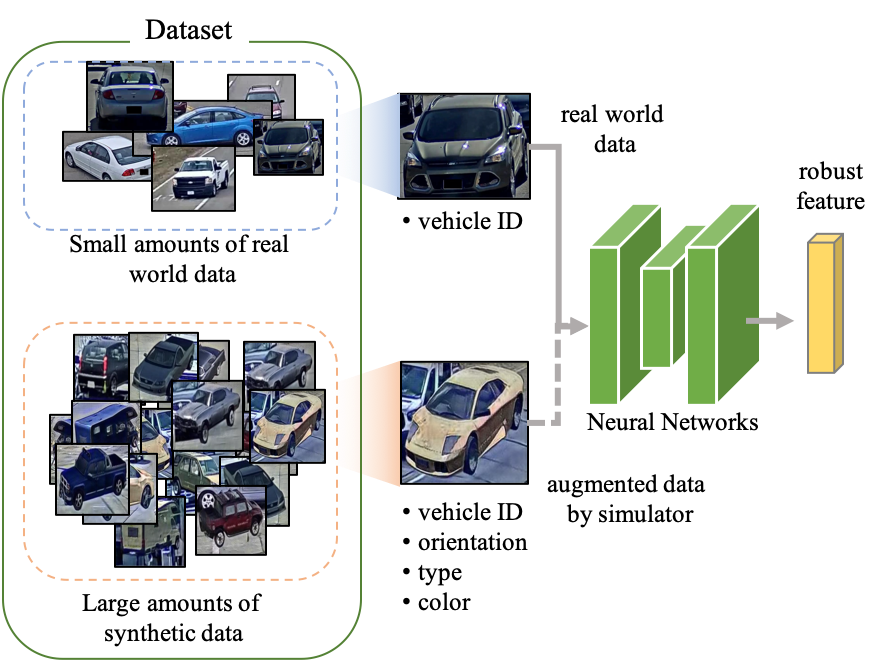}
\end{center}
  \caption{Proposed synthetic-to-real domain adaptation method to improve vehicle re-identification performance. It can be difficult to obtain meaningful labels for real data, but it is relatively simple for synthetic data.}
\end{figure}

Deep learning methods \cite{Tan2019MulticameraVT, Huang2019MultiViewVR, Lv2019VehicleRW} are commonly employed to tackle this complex vehicle Re-ID task with significant progress. These models extract features using deep learning networks and distinguish vehicles by comparing feature distances. However, they require large datasets for training and improved performance, which rapidly becomes a drawback. Many studies [30] have confirmed that more training data provides better model performance. Therefore, data from real environments require considerable annotation workload. On the other hand, domain adaptation approaches employ inexpensive synthetic data to replace real data.

This paper explores how to improve model performance using inexpensive synthetic data (see Fig. \ref{fig:fig1}). We adopted an adversarial domain adaptation approach \cite{Ganin2014UnsupervisedDA} where an artificial neural network (ANN) learns the best discriminating features for classification using real data, while simultaneously learning indistinguishable features between real and synthetic data \cite{Ajakan2014DomainAdversarialNN} \cite{Ganin2016DomainAdversarialTO}. To implement this concept, we introduce a domain discrimination layer and associated cross-entropy loss to train the whole network to be indiscriminative for both domains. We also adopted a semi-supervised learning method to better exploit specific synthetic data labels, such as color, type, and orientation. Since these labels only exist for synthetic data, a semi-supervised learning approach that can handle unlabeled data is appropriate to improve performance. In training, classification losses for the exclusive labels are selectively applied depending on the data domain \cite{Zhai2019S4LSS}. The proposed model trained on real and synthetic data from the AI City Challenge using domain adaptation and semi-supervised learning approaches achieved 12.9\% improvement over the baseline model, which was trained with only real data.

This work proposes a novel synthetic-to-real domain adaptation network StRDAN framework, with major contributions as follows.

\begin{itemize}
\item StRDAN can be successfully trained with inexpensive large-scale synthetic as well as real data to improve performance.
\item We propose a new training approach for StRDAN, combining domain adaptation and semi-supervised learning methods and corresponding losses.
\item StRDAN shows significant improvement over the baseline model for two significant data sets: VeRi \cite{liu2016deep} and CityFlow-ReID \cite{Tang2019CityFlowAC}.
\end{itemize}

\section{Related Work}
This section reviews relevant prior studies regarding vehicle Re-ID and domain adaptation methods with synthetic data.

{\bf Vehicle Re-ID: } 
Vehicle Re-ID methods generally incorporate contrastive loss and spatio-temporal features. Previous studies \cite{Liu2016LargescaleVR, liu2016deep, Liu2018PROVIDPA} have proposed several contrastive loss based methods, such as siamese networks, triplet loss, and metric learning. Liu \etal \cite{liu2016deep} introduced the VeRi dataset, being the first large scale vehicle Re-ID benchmark. Spatio-temporal features are critical for performance improvement and have helped vehicle Re-ID studies to achieve huge progress. Tan \etal \cite{Tan2019MulticameraVT} used spatial-temporal features for multi-camera vehicle tracking and vehicle Re-ID to win the AI City Challenge in 2019 \cite{naphade20192019}. Shen \etal \cite{shen2017learning} proposed a two-stage framework to match visual appearance based on an long-short term memory (LSTM) based path inference mechanism. 

{\bf Domain Adaptation with Synthetic Data: } 
To overcome the lack of data, Zhou \etal \cite{zhou2018aware,Zhou2018VehicleRB} proposed a method to improve Re-ID performance by augmenting various viewpoint vehicle images with generative adversarial networks (GANs). Performance significantly reduces when deploying trained models onto new datasets due to differences among the datasets, commonly called domain bias. Peng \etal \cite{peng2019cross} proposed a domain adaptation framework to address this problem, incorporating an image-to-image translation network and an attention based feature learning network. The VehicleX \cite{Yao19VehicleX} simulator also leverages synthetic data and domain randomization to overcome the reality gap \cite{tobin2017domain, tremblay2018training}. Although Liu \etal \cite{liu2019supervised} proposed a domain adaptation method, they only considered real-to-real domain adaptation. The recently proposed PAMTRI approach \cite{Tang2019PAMTRIPM} uses synthetic data to improve model performance and has similar architecture to the proposed StRDAN framework. However, PAMTRI requires considerable effort to obtain vehicle pose and labels for real data, whereas StRDAN uses domain adaptation to utilize synthetic data and semi-supervised learning does not require additional annotation workload. Thus, StRDAN is somewhat simpler and easier to train.

\begin{figure*}[t!]
\begin{center}
\includegraphics[width=17.0cm]{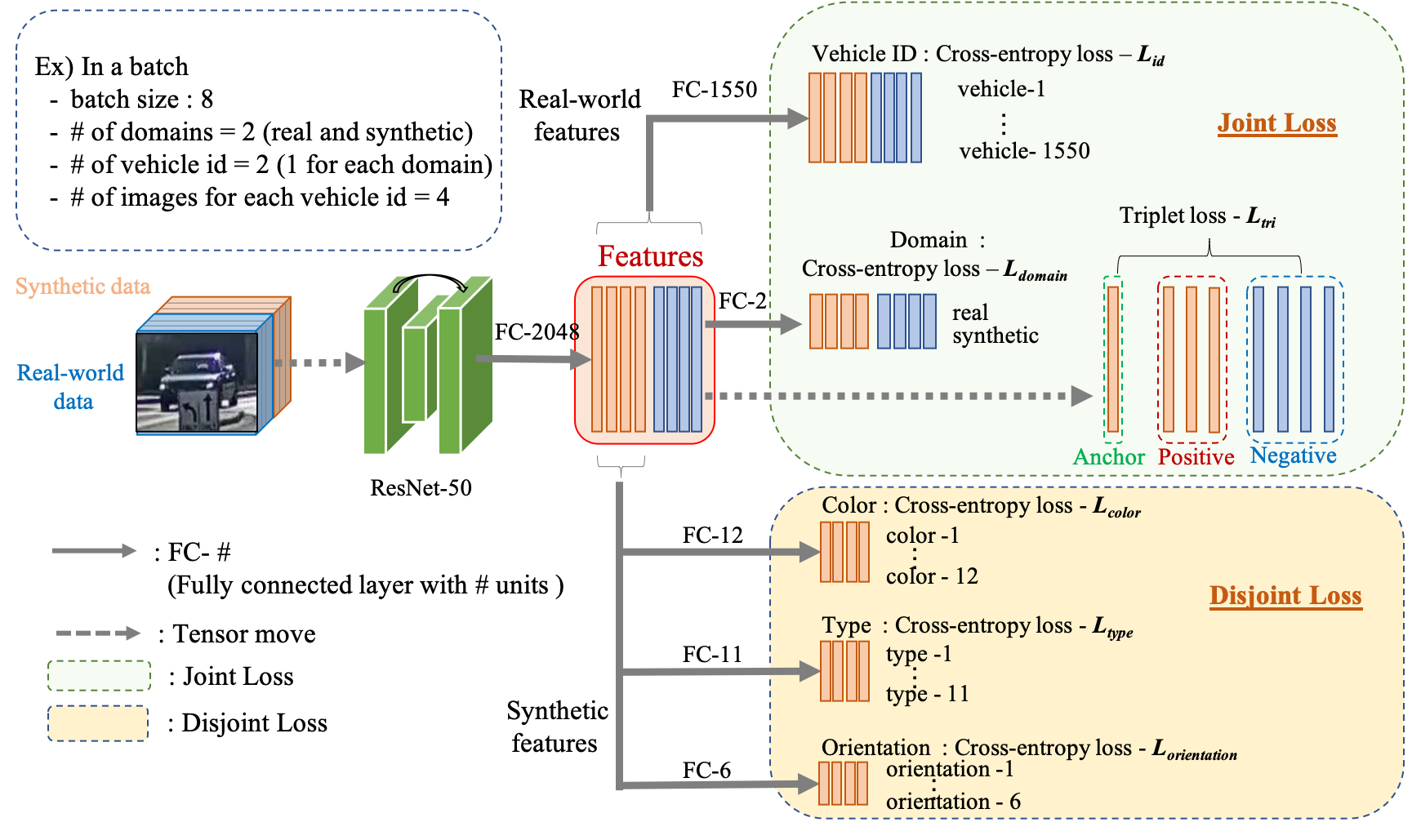}
\end{center}
  \caption{Proposed synthetic-to-real domain adaptation network architecture comprising a ResNet-50 backbone for feature extraction and five fully connected softmax layers for classification, trained using joint and disjoint losses between synthetic and real data.}
\label{fig:short}
\end{figure*}

\section{Proposed Synthetic-to-Real Domain Adaptation Network}
\subsection{Datasets}
We developed an ANN using real and synthetic vehicle datasets provided for Track 2 of the 2020 AI City Challenge. The real dataset was the CityFlow-reID dataset, a subset of CityFlow made available for the Track 2 challenge comprising 56,277 images for 666 unique vehicles collected from 40 cameras, with 36,935 images from 333 vehicle identities for training, and 18,290 images from the other 333 identities for testing. The remaining 1052 images in the test set were provided as query data.

The synthetic vehicle dataset comprised 192,150 images from 1,362 distinct vehicles created using the VehicleX \cite{Yao19VehicleX} synthetic dataset generator, forming an augmented training set. The synthetic dataset includes vehicle ID, color, type, and object orientation; whereas the real dataset includes only vehicle ID. Vehicles were distinguished into 12 colors and 11 types, and orientation was represented by rotation angle [0, 360) on the horizontal plane.

We trained and evaluated the proposed StRDAN model using the VeRi real dataset [8] and City Challenge synthetic data to examine validity and robustness for the approach. The Veri dataset contains over 50,000 images for 776 vehicles captured by 20 cameras, with the training set containing 37,781 images for 576 vehicles, and the testing set containing 11,579 images for 200 vehicles. The VeRi dataset includes labels for vehicle color and type.

\subsection{Overall Architecture}
Figure \ref{fig:short} shows the proposed overall StRDAN architecture. The model comprises a backbone network for feature extraction and multiple fully connected (FC) softmax layers for classification. Input images are batch sampled in equal numbers from the real and synthetic datasets. For a mini-batch, \textit{n} different vehicle identities are chosen from the real and synthetic datasets, respectively, then \textit{m} samples are randomly selected from these chosen images. Therefore, each batch contains $2 \times n \times m$ images.

The backbone network extracts a highly abstracted feature vector (\textit{dim} = 2048) from the input image. In principle, any convolution neural network (CNN) designed for image classification can be used as the backbone network, and a variety of CNNs have been employed in previous studies, including VGG-CNN-M1024 \cite{Chatfield2014ReturnOT}, MobileNet \cite{Howard2017MobileNetsEC}, and ResNet \cite{He2016IdentityMI}, as vehicle Re-ID model backbone. We selectedResNet-50 as the backbone network for StRDAN. Feature maps extracted by the backbone network are flattened and fed into various FC softmax layers to classify vehicle id, real or synthetic, color, type, and orientation. Outputs are then fed into five cross-entropy loss functions and one triplet loss function. StRDAN was end-to-end trained by updating the network parameters to reduce total loss, combining cross-entropy and triplet losses.

\subsection{Key Features}
\textbf{Adversarial Domain Adaptation}. 
An annotated dataset is essential for deep neural network supervised learning. However, collecting and manually annotating large datasets is time consuming and expensive. Therefore, the VehicleX approach was introduced in the AI City Challenge to generate automatically labeled data using a graphic simulator and hence overcome the dearth of real data. However, synthetic data has similar but different distributions than real data. Therefore, it is necessary to train the ANN to predict the classification, regardless of the input domain.

We adopted the adversarial domain adaptation approach, where the ANN learns features that are most discriminative for classification on the real domain and simultaneously as indistinguishable as possible between the real and synthetic domains \cite{Ajakan2014DomainAdversarialNN} \cite{Ganin2016DomainAdversarialTO}. To implement this approach, we introduced a domain discrimination layer and its associated cross-entropy loss train the network to be indiscriminate to the domains. We also introduced a vehicle-id classification layer and its associated cross-entropy loss along with triplet loss to train the network to better discriminate vehicle identities and shape signatures.

\textbf{Semi-supervised Learning}.
In contrast with the real data, synthetic data includes vehicle type, color, and orientation labels, and we use these labels under multitask learning to improve generalization performance for all tasks \cite{Zhang2017ASO}. Many semi-supervised learning approaches improve learning accuracy by combining a small amount of labeled data with a large amount of unlabeled data during training. Zhai \etal?™s work \cite{Zhai2019S4LSS} created artificial labels for unlabeled and labeled data and utilize them in training, and this approach inspired us to use joint and disjoint labels between real and synthetic data to improving performance. Joint labels attached to real and synthetic data are vehicle ID and domain (real or synthetic), whereas disjoint labels were attached to only synthetic data and includes vehicle type, color, and orientation. Losses were classified as joint or disjoint losses, associated with joint and disjoint labels, respectively, as shown in Fig. \ref{fig:short}. Triplet loss is classified as a joint loss because the vehicle id contributes to distinguishing batch images into anchor, positive, or negative images.

The semi-supervised learning approach considered here has learning objective 
\begin{equation}
    \operatorname*{min}_\theta\; \mathcal{L}_{joint}(\theta) + w\mathcal{L}_{disjoint}(\theta),
\end{equation}
where $\mathcal{L}_{joint}$ is the joint loss defined in both real and synthetic domains and $\mathcal{L}_{disjoint}$ is the disjoint loss defined in the synthetic domain. $\theta$ is parameters of the network. In the next section, we will describe the losses in more detail.

\section{Loss Function}
\subsection{Joint Losses}
\textbf{Vehicle ID}.
Cross-entropy following the softmax function is the most commonly employed loss for image classification, and can be represented for vehicle ID classifier, Lid, as follows
\begin{equation}
L_{id} = - \frac{1}{N}\sum_{i=1}^{N}\sum_{j=1}^{C} y_{ij} \log(\hat{y}_{ij})
\end{equation}
where $N$ denotes the number of images in a mini-batch, $C$ is the number of classes, $y_{ij}$ is the $j^{\mathit{th}}$ element of an one hot encoded vector for the ground-truth of the $i^{\mathit{th}}$ sample in a mini-batch, and $\hat{y}_{ij}$ is the $j^{\mathit{th}}$ element of the softmax FC layer for the $i^{\mathit{th}}$ image. 

\textbf{Domain}.
We adopted the adversarial domain adaptation approach, with real and synthetic domains. A softmax FC layer was added to the backbone network for domain discrimination, with loss function to train the network to be indiscriminate to the domains,
\begin{equation}
L_{domain} = \frac{1}{N}\sum_{i=1}^{N} y_i  \log(\hat{y}_i) + (1 - y_i) \log(1 - \hat{y}_i).
\end{equation}
Domain discrimination loss is defined as the negative of binary cross-entropy loss. Since cross-entropy loss trains the network to discriminative between the domains, the negative loss trains the model to be less discriminating. Thus, if a vehicle captured by a camera is drawn by a graphic simulator in the same orientation, features extracted from a synthetic image would be similar to that from a real image since domain dependent features are suppressed. The negative cross-entropy loss function was implemented by a gradient reversal layer \cite{Ganin2014UnsupervisedDA}.

\textbf{Triplet Loss}. 
In a mini-batch that contains $P$ identities and $Q$ images for each identity, each image (anchor) has $Q - 1$ images of the same identity (positives) and $(P-1) \times Q$ images of different identities (negatives). Triplet loss pulls the positive pair ($a$, $p$) together while pushing the negative pair ($a$, $n$) away by some margin. Thus, triplet loss trains the network to minimize the distance between features from the same image classes and simultaneously maximizes distance between features from different image classes. Triplet loss is defined as \cite{Hermans2017InDO} 

\begin{equation} \label{eq1}
\begin{split}
L_{tri} = \sum_{i=1}^{P}\sum_{a=1}^{Q}\Bigg[m &+ \operatorname*{max}_{\substack{p=1\dots Q\\p\neq a}}D(v_{a, i}, v_{p, i})\\
& - \operatorname*{min}_{\substack{j=1\dots P\\n=1\dots Q\\j\neq i}}D(v_{a, i}, v_{n,j})\Bigg]_{+}
\end{split}
\end{equation}
where $v_{a, i}$ is the prediction vector for the $a^{th}$ image of the $i^{th}$ identity group, and $m$ is the margin to control the difference between positive and negative pair distances and helps cluster the distribution more densely.


\subsection{Disjoint Losses}

\textbf{Color, Type, and Orientation}. Softmax cross-entropy loss was applied for these three targets. Orientation is continuous and numerical, whereas color and type are categorical and nominal. Therefore, it seems reasonable to use regression to predict orientation, but orientation is a difficult problem for regression due to the wide range of the regression target. Indeed, experiments showed that optimization would not converge for regression. Therefore, we predicted orientation as direct classification into n discrete bins, with softmax cross-entropy loss \cite{Zhou2018OnTC} or \cite{Gidaris2018UnsupervisedRL}, dividing the 360 degrees of orientation space into six bins of 60 degrees. Cross-entropy losses for color, type, and orientation were only applied to synthetic images and set to zero for real images. The loss function can be expressed as 
\begin{equation}
L_{x} = - \frac{1}{N}\sum_{i=1}^{N} \sum_{j=1}^{C} \delta_i y_{ij} \log(\hat{y}_{ij}) , \; \delta_i \in \{1, 0\},
\end{equation}
where $x$ is one of color, type, and orientation, and $\delta_i$ is a mask value that is set to 1 if the $i^{th}$ data in a mini-batch has $x$, and 0 otherwise.

\section{Experiments}
\subsection{Evaluation Metric}
We used rank-K mean Average Precision (mAP), the official AI City Challenge evaluation metric, to evaluate model performances. mAP measures the mean average precision for each query considering only the top K matches, where we chose K = 100. Average precision was computed for each query image from the area under the precision-recall curve, and then the mean of the average precision over all queries was computed.

\subsection{Implementation}
The chosen backbone network, ResNet-50, was initialized with weights pre-trained on ImageNet \cite{Russakovsky2015ImageNetLS} to accelerate training. We trained the model end-to-end with an AMSGrad optimizer \cite{reddi2019convergence} for 60 epochs. Initial learning rate = 0.0003, reduced by 0.1 after 20 and 40 epochs. Weight decay factor for L2 regulation was set = 0.0005, and batch size = 64. For each mini-batch, two different vehicle-IDs were selected from each of the real and synthetic datasets, and four images with the same ID were sampled. Therefore, 16 different images with four different IDs from real and synthetic datasets were sampled. Input images were resized to 128 x 256 pixel, and we employed horizontal flip and random erasure augmentations. Postprocessing used the re-ranking algorithm  \cite{Zhong2017RerankingPR}, to order the distance matrix between features with Jaccard and original output distance.

\begin{table}
\caption{Evaluation results of the StRDAN trained with CityFlow-reID and VehicleX datasets for the 2020 AI City Challenge, Track 2. The results are from the official evaluation leaderboard of the challenge. }
\label{table:ai}
\begin{center}
\begin{tabular}{c|c|c|c|c|c|c||c}
\hline
Case & O & C & T & V & D & Dataset & mAP\\
\hline
1& & & &\checkmark& &R& 25.5\\
\hline
2&\checkmark& & &\checkmark&\checkmark&R+S& No Conv.\\
\hline
3& &\checkmark& &\checkmark&\checkmark&R+S&  35.2\\
\hline
4& & &\checkmark&\checkmark&\checkmark&R+S&  \textbf{38.4}\\
\hline
5&\checkmark&\checkmark& &\checkmark&\checkmark&R+S&  34.1\\
\hline
6&\checkmark& &\checkmark&\checkmark&\checkmark&R+S&  37.5\\
\hline
7& &\checkmark&\checkmark&\checkmark&\checkmark&R+S&  35.3\\
\hline
8&\checkmark&\checkmark&\checkmark&\checkmark&\checkmark&R+S&  34.0\\
\hline
\end{tabular}
\end{center}
\end{table}

\begin{table}
\caption{Evaluation results of the StRDAN trained with VeRi and VehicleX datasets.}
\label{table:veri}
\begin{center}
\begin{tabular}{c|c|c|c|c|c|c||c}
\hline
\cline{1-8}
Case & O & C & T & V & D & Dataset & mAP\\
\hline
1& & & &\checkmark& & R &73.0\\
\hline
2&\checkmark& & &\checkmark&\checkmark&R+S&\textbf{76.1}\\
\hline
3& &\checkmark& &\checkmark&\checkmark&R+S&74.2\\
\hline
4& & &\checkmark&\checkmark&\checkmark&R+S&74.9\\
\hline
5&\checkmark&\checkmark& &\checkmark&\checkmark&R+S&74.7\\
\hline
6&\checkmark& &\checkmark&\checkmark&\checkmark&R+S&75.3\\
\hline
7& &\checkmark&\checkmark&\checkmark&\checkmark&R+S&74.8\\
\hline
8&\checkmark&\checkmark&\checkmark&\checkmark&\checkmark&R+S&74.6\\
\hline
\end{tabular}
\end{center}

\textbf{Notes}: O, C, T, V, D = orientation, color, type, vehicle ID, and domain, respectively.\\
R, S = real and synthetic data, respectively.\\
No Conv. = no convergence\\
mAP = mean average precision\\
Boxes are checked if target loss was included. \\
Best result is shown in bold.
\end{table}

\begin{table}
\caption{Comparing Deep learning methods on the VeRi dataset.}
\label{table:verio}
\begin{center}
\begin{tabular}{c||c}
\hline
\cline{1-2}
Method & mAP\\
\hline
FACT\cite{Liu2016DeepRD} & 18.7\\
\hline
ABLN\cite{Zhou2018VehicleRB} & 24.9\\
\hline
OIFE\cite{Wang2017OrientationIF} & 48.0\\
\hline
PROVID\cite{Liu2018PROVIDPA} & 48.5\\
\hline
PathLSTM\cite{shen2017learning} & 58.3\\
\hline
GSTE\cite{Bai2018GroupSensitiveTE} & 59.5\\
\hline
VAMI\cite{Zhou2018ViewpointAwareAM} & 61.3\\
\hline
BA\cite{Kumar2019VehicleRA} & 66.9\\
\hline
BS\cite{Kumar2019VehicleRA} & 67.6\\
\hline
PAMTRI\cite{Tang2019PAMTRIPM} & 71.9\\
\hline
StRDAN (R, baseline) & 73.0\\
\hline
StRDAN (R+S, best) & \textbf{76.1}\\
\hline
\end{tabular}
\end{center}
\end{table}

\subsection{Results and Discussion}
We trained and evaluated our models using the CityFlow-reID and VeRi real datasets along with synthetic data generated by VehicleX, with selected disjoint losses as shown in Table \ref{table:ai} and Table \ref{table:veri}.

\textbf{Performance on AI City Dataset}. 
The baseline (Case 1) model comprised the backbone network and vehicle ID classifier. The baseline was trained with the real dataset using vehicle-ID cross-entropy and triplet losses. Table \ref{table:ai} shows that the proposed domain adaptation and semi-supervised learning approaches significantly improved model performance compared with the baseline by at least 8.5\% (Case 8) up to 12.9\% (Case 4). The proposed model exhibited best performance for Case 4, where only vehicle type was considered, whereas Case 8 considered all three labels and exhibited the worst performance.

\textbf{Performance on VeRi Dataset}.
Table \ref{table:veri} also shows that domain adaptation and semi-supervised learning approaches with synthetic dataset and additional losses helped improve performance by at least 1.2\% (Case 3) up to 3.1\% (Case 2). In contrast to the AI City dataset, the best performance was when considering only orientation (Case 2). However, the model could not converge with the AI City dataset. Veri data model performances were much superior than those for AI City data. 
Table \ref{table:verio} compares the proposed StRDAN approach with other methods. All models were trained using only the VeRi dataset, aside from PAMTRI and StRDAN (R+S). The proposed StRDAN (R+S) model outperformed all other considered methods.

\textbf{Domain Adaptation and Semi-supervised Learning}. 
The experimental results verified that domain adaptation and semi-supervised learning approaches help extract more important semantic features for vehicle Re-ID. However, further research is required regarding unexpected phenomena as follows.

\begin{itemize}
\item The best model was trained with only one loss of three disjoint losses.
\item Performance degraded with including more disjoint losses. 
\item Best performance depended on the real dataset.
\end{itemize}

\section{Conclusions}
This paper proposes using domain adaptation and semi-supervised learning to fully utilize synthetic data. Experiment results confirmed that increasing training data via with domain adaptation improved performance. We also showed that using semi-supervised learning with labels only available for synthetic data helped the model extract more semantic features. 

Future work will investigate the following issues.
\begin{itemize}

\item Synergy between disjoint losses and real-world data dependency on disjoint losses, as discussed above.

\item Reality effects on synthetic data. Image data synthesized by VehicleX was far from realistic and hence easily distinguishable from real image data. More realistic synthetic data from more sophisticated simulation software could further improve performance.

\item Orientation prediction. We converted orientation regression to six bin classification, but we have not optimized bin count. Since orientation is a key feature to identify vehicles captured at various camera angles, proper orientation representation will also help to improve performance.

\end{itemize}

{\small
\bibliographystyle{ieee_fullname}
\bibliography{egbib}
}

\end{document}